\documentclass[a4paper,fleqn]{cas-sc}

\usepackage[numbers]{natbib}

\usepackage{enumerate}
\usepackage{enumitem}
\setlist[enumerate,1]{label=\arabic*.,font=\textup}

\usepackage{amsthm}

\usepackage{algorithm}
\usepackage{algpseudocode}

\def\tsc#1{\csdef{#1}{\textsc{\lowercase{#1}}\xspace}}
\tsc{WGM}
\tsc{QE}
\tsc{EP}
\tsc{PMS}
\tsc{BEC}
\tsc{DE}

\begin{document}
\let\WriteBookmarks\relax
\def\floatpagepagefraction{1}
\def\textpagefraction{.001}

\shorttitle{HGRL-DTA}

\shortauthors{Zhaoyang Chu et~al.}

\title [mode = title]{Hierarchical Graph Representation Learning for the Prediction of Drug-Target Binding Affinity}

\author[1]{Zhaoyang Chu}[style=chinese, orcid=0000-0003-4333-8063]

\author[1]{Shichao Liu}[style=chinese]
\cormark[1]

\author[1]{Wen Zhang}[style=chinese]
\cormark[1]

\address[1]{College of Informatics, Huazhong Agricultural University, Wuhan, 430070, China}

\cortext[cor1]{Corresponding author: zhangwen@mail.hzau.edu.cn, scliu@mail.hzau.edu.cn}

\begin{abstract}
The identification of drug-target binding affinity (DTA) has attracted increasing attention in the drug discovery process due to the more specific interpretation than binary interaction prediction. Recently, numerous deep learning-based computational methods have been proposed to predict the binding affinities between drugs and targets benefiting from their satisfactory performance. However, the previous works mainly focus on encoding biological features and chemical structures of drugs and targets, with a lack of exploiting the essential topological information from the drug-target affinity network. In this paper, we propose a novel hierarchical graph representation learning model for the drug-target binding affinity prediction, namely HGRL-DTA. The main contribution of our model is to establish a hierarchical graph learning architecture to incorporate the intrinsic properties of drug/target molecules and the topological affinities of drug-target pairs. In this architecture, we adopt a message broadcasting mechanism to integrate the hierarchical representations learned from the global-level affinity graph and the local-level molecular graph. Besides, we design a similarity-based embedding map to solve the cold start problem of inferring representations for unseen drugs and targets. Comprehensive experimental results under different scenarios indicate that HGRL-DTA significantly outperforms the state-of-the-art models and shows better model generalization among all the scenarios.
\end{abstract}


\begin{highlights}
\item We propose a novel hierarchical graph representation learning model for drug-target binding affinity prediction, named HGRL-DTA. The model can capture global-level affinity relationships and local-level chemical structures involving drug/target molecules synergistically, and integrate such hierarchical information with a message broadcasting strategy.
\item To solve the cold start problem of inferring representations for unseen drugs and targets, we design a similarity-based embedding map from known to unknown drugs/targets, which can infer the representation of the unknown drug (target) by aggregating the representations of its most similar known drugs (targets).
\item Extensive experiments under four experimental scenarios are conducted to evaluate the performance of the proposed model. Compared with four state-of-the-art methods, HGRL-DTA achieves significantly better model generalization among all four scenarios.
\end{highlights}


\maketitle

\section{Introduction}
Molecular drugs exert their therapeutic effects by working as ligands to interact with target proteins and activate or inhibit the biological process of targets. Investigation of drug-target interactions (DTI) plays a crucial role in drug discovery, which helps in understanding the specific interactions between drug compounds and target proteins. However, due to the numerous possible protein-compound combinations with more than 5,000 potential proteins \cite{Gilson2015} and over 100 million drug candidate compounds \cite{Kim2019}, computational approaches have become increasingly necessary to predict drug-target interactions. The existing approaches for drug-target interaction prediction mainly fall into two categories. The first category of methods formulates the interaction prediction as a binary classification task (interacts or not) \cite{Chen2020a, Gao2018b, Huang2020, Lee2019, Tsubaki2018}. However, these methods suffer from two primary defects: (1) the inability to differentiate between true-negative and unknown interactions, and (2) the binary relationship is unable to indicate the continuous binding affinity that quantifies how tightly a drug binds to a target \cite{Pahikkala2014}. To tackle the above limitations, the other kind of approaches considers the interaction prediction as a regression problem to predict continuous binding affinities, which has attracted more and more attention in recent studies \cite{Ballester2010, Jiang2020, Li2021, Ozturk2018, Ru2022}.

Most computational approaches for binding affinity prediction focus on the usage of 3D structures of protein-compound complexes \cite{Ballester2010, Ganotra2018, Jimenez2018, Li2021, Ragoza2017}. Although these 3D structure-based methods have achieved relatively high predictive performance, they remain limitations because the exploitation of 3D structural data is costly and time-consuming. In practice, the co-crystallized 3D structures of protein-compound complexes are usually difficult to obtain, and predicting them by docking individual structures of proteins and compounds together also remains a challenging task \cite{Karimi2019, Karimi2021}. For this reason, structure-free prediction of drug-target binding affinities has emerged to overcome the limitations of 3D structure-based methods, without the analysis of 3D structural data and the molecular docking process \cite{Thafar2019, Wang2021b}. The early structure-free approaches utilize statistic machine learning to predict binding affinities of drug-target pairs with hand-crafted features \cite{He2017}, similarity information \cite{Pahikkala2014} of drugs and targets. However, these approaches heavily rely on expert knowledge and feature engineering, which may lead to limited accuracy and generalization of models.

Recently, deep learning \cite{Lecun2015} has achieved remarkable success in various machine learning tasks and demonstrated its expansion capability for diverse application scenarios. Deep learning-based affinity prediction has become a popular research \cite{Ozturk2018}, which can automatically learn feature representations of drugs and targets in an end-to-end way and show its superior performance. Most of the existing works represent drug compounds as simplified molecular input line entry system (SMILES) strings and target proteins as amino acid sequences, respectively, and learn hidden patterns from the sequence data using various neural networks, such as convolutional neural networks (CNNs) \cite{Abbasi2020, Karimi2019, Ozturk2018, Zhao2019}, recurrent neural networks (RNNs) \cite{Abbasi2020, Karimi2019} and generative adversarial networks (GANs) \cite{Zhao2020}. Moreover, owing to the significant advance of deep learning on graph-structured data \cite{Gao2018a, Sun2019}, some works have been designed to model 2D graph structure information of drug compounds \cite{Jiang2020, Karimi2019, Lin2020, Nguyen2020} and target proteins \cite{Jiang2020, Nguyen2021}.

Despite the success of representation learning for drugs and targets, the affinity information cannot be properly exploited by the previous deep learning-based methods, which only use affinities as the true labels for model optimization. Recent studies aim to construct heterogeneous networks based on binary drug-target affinities \cite{Cheng2021, Luo2017, Peng2021, Sun2020, Wan2019}, which tend to lose more realistic information hidden in continuous affinities. To take advantage of the essential topological information deriving from drug-target affinity relationships, the widely used graph representation learning method, i.e., graph neural networks (GNNs) \cite{Wu2020}, can be applied to facilitate the predictive performance of models. However, the graph representation learning method works based on the graph connectivity of the affinity data, which cannot learn representations for unseen drugs/targets. In addition, previous graph-based binding affinity prediction works only concentrate on utilizing the chemical structures to learn representations for drugs and targets \cite{Jiang2020, Nguyen2020}. The combined analysis of affinity relationships and chemical structures is often overlooked by these methods, which is essentially necessary and proven effective in some relative researches \cite{Bai2020, Wang2021a, Wang2021c}.

To overcome the limitations mentioned above, we propose a novel hierarchical graph representation learning method for the drug-target binding affinity prediction, namely HGRL-DTA, which can capture global-level affinity relationships and local-level chemical structures involving drug/target molecules synergistically. HGRL-DTA establishes a hierarchical graph architecture, where drug-target affinity relationships are encoded as a global-level affinity graph and each node inside it, i.e., drug or target, is encoded as a local-level molecular graph. We firstly utilize a popular graph neural network, i.e., graph convolutional network (GCN) \cite{Kipf2017}, to encode affinity relationships and chemical structures of drugs and targets. Then, we design a message broadcasting mechanism to integrate the hierarchical information learned from the global-level affinity graph and local-level molecular graph. After that, we reuse GCNs to conduct a refinement process for molecular representations and readout the final embeddings of drugs and targets to make drug-target binding affinity predictions. Moreover, to solve the cold start problem of inferring representations for unseen drugs and targets, we build a similarity-based embedding map from known to unknown drugs/targets, which can infer the unknown drug (target) by aggregating the representations of its most similar known drugs (targets).

\section{Preliminaries}
\label{sec:preliminaries}
This section introduces some definitions used in our proposed method and describes the problem formulation of drug-target binding affinity prediction.\vspace{5pt}

\textit{Definition 2.1.} \textbf{Affinity Graph.} Given a set of drugs $\mathcal{V}_d = \{d_1, \dots, d_{n_d}\}$ and a set of targets $\mathcal{V}_t = \{t_1, \dots, t_{n_t}\}$, we define the affinity graph as a weighted bipartite graph $\mathcal{N} = \{\mathcal{V}_d, \mathcal{V}_t, \mathcal{E}_a\}$, where $\mathcal{E}_a \subseteq \mathcal{V}_d \times \mathcal{V}_t$ denotes the set of drug-target pairs with known affinities. Note that a subset of $\mathcal{E}_a$ is selected as the training set $\mathcal{T}$, and the rest of samples as the test set are masked during training.\vspace{5pt}

Specifically, the drug-target pair set $\mathcal{E}_a$ can be formulated as a drug-target co-occurrence matrix $\mathbf{Y} \in \mathbb{R}^{n_d \times n_t}$, whose entries represents the continuous affinity values of the respective drug-target pairs. Further, we formulate the affinity graph $\mathcal{N}$ as a normalized adjacency matrix $\mathbf{A} \in [0, 1]^{(n_d + n_t) \times (n_d + n_t)}$, where $\mathbf{A}_{i, j + n_d} = f_{norm} (\mathbf{Y}_{i, j})$ if there exists a known affinity from the $i$-th drug to the $j$-th target (i.e., $(d_i, t_j) \in \mathcal{E}_a$) and otherwise $\mathbf{A}_{i, j} = 0$, $f_{norm} (\cdot)$ denotes the normalization function. Note that $\mathbf{A}$ is symmetric because $\mathcal{N}$ is an undirected graph. The initial signals of nodes (i.e., drugs and targets) are summarized in a matrix $\mathbf{X} \in \mathbb{R}^{(n_d + n_t) \times d_{ag}}$, where $d_{ag}$ denotes the signal dimension.\vspace{5pt}

\textit{Definition 2.2.} \textbf{Drug Molecular Graph.} Given a drug $d_i$ and a set of atoms $\mathcal{V}_a = \{a_1, \dots, a_{n_a}\}$ inside it, the drug molecular graph is defined as $\mathcal{G}_{d_i} = \{\mathcal{V}_a, \mathcal{E}_b\}$, where $\mathcal{E}_b \subseteq \mathcal{V}_a \times \mathcal{V}_a$ denotes the set of atom-atom covalent bonds. An initial feature vector $\mathbf{x}_{a_u} \in \mathbb{R}^{d_{dg}}$ is assigned for each atom $a_u$, where $d_{dg}$ denotes the feature dimension.\vspace{5pt}

\textit{Definition 2.3.} \textbf{Target Molecular Graph.} Given a target $t_j$ and a set of residues $\mathcal{V}_r = \{r_1, \dots, r_{n_r}\}$ inside it, the target molecular graph is defined as $\mathcal{G}_{t_j} = \{\mathcal{V}_r, \mathcal{E}_c\}$, where $\mathcal{E}_c \subseteq \mathcal{V}_r \times \mathcal{V}_r$ denotes the set of residue-residue contacts. Each residue $r_p$ is encoded as an attributed vector $\mathbf{x}_{r_p} \in \mathbb{R}^{d_{tg}}$, where $d_{tg}$ is the attribution dimension.\vspace{5pt}

\textit{Definition 2.4.} \textbf{Drug-Target Binding Affinity Prediction.} Given the training set $\mathcal{T}$, the affinity graph $\mathcal{N}$, the drug molecular graphs $\{\mathcal{G}_{d_i}\}_{i=1}^{n_d}$ and the target molecular graphs $\{\mathcal{G}_{t_j}\}_{j=1}^{n_t}$, our goal of drug-target binding affinity prediction is to learn a mapping function $\mathit{\Theta}(\omega) : (\mathcal{N}, \mathcal{G}_{d_i}, \mathcal{G}_{t_j}) \to y_{i,j}$ to precisely predict the binding affinity $y$, where $\omega$ is the trainable parameter and $(d_i, t_j) \in \mathcal{T}$.

\section{Model framework}
\begin{figure}[htp]
    \centering
    \includegraphics[width=1\textwidth]{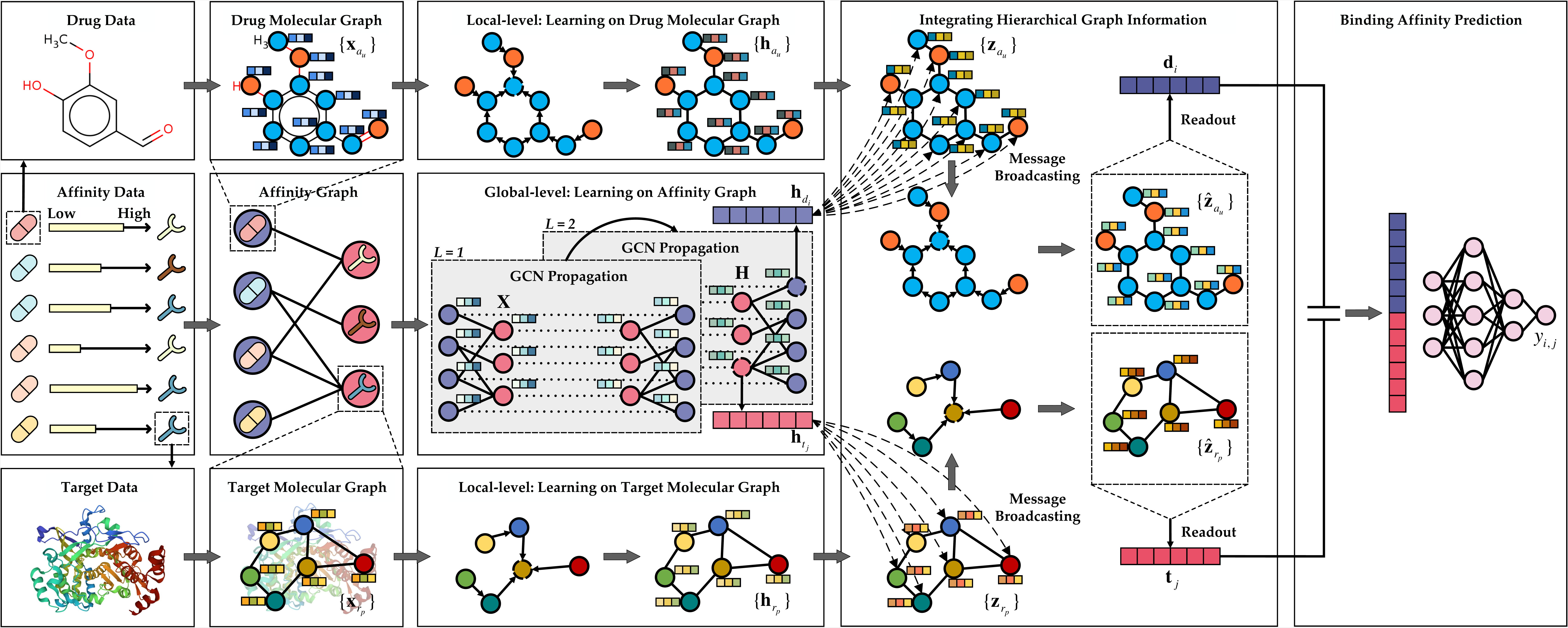}
    \caption{Overview of HGRL-DTA.}
    \label{fig:Framework}
\end{figure}

In this section, we introduce the proposed hierarchical graph representation learning model for the drug-target binding affinity prediction, namely HGRL-DTA. The model builds a hierarchical graph setting, where the affinity data is formulated as an affinity graph and each node inside it, i.e., drug or target, is formulated as a molecular graph. We firstly utilize GCNs to learn the global-level affinity relationships on the affinity graph and the local-level chemical structures on the molecular graph, respectively. Then, we integrate the learned hierarchical graph information using a message broadcasting mechanism and reuse GCNs to refine the molecular representations. Finally, we readout the representations of drugs and targets and combine them to make binding affinity predictions. Figure \ref{fig:Framework} illustrates the framework of our proposed HGRL-DTA model.

\subsection{Global-level graph representation learning on affinity graph}
In this paper, we build a graph-based encoder with a two-layer GCN to capture the global-level affinity relationships. This encoder learns global-level representations for drugs and targets through neighbourhood aggregation and feature transformation.{

Given the training set $\mathcal{T}$ and the affinity graph $\mathcal{N} = \{\mathcal{V}_d, \mathcal{V}_t, \mathcal{E}_a\}$, we recompose the drug-target pair set $\mathcal{E}_a$ by removing test pairs, i.e., masking test entries in the drug-target matrix $\mathbf{Y}$. Then, we preprocess the adjacency matrix $\bf{A}$ deriving from $\mathbf{Y}$:
\begin{equation}
\mathbf{\hat{A}} = \mathbf{D}^{-\frac12} \mathbf{A} \mathbf{D}^{-\frac12}
\label{equ:adj_preprocess}
\end{equation}
where $\mathbf{D} \in \mathbb{R}^{(n_d + n_t) \times (n_d + n_t)}$ denotes a diagonal matrix whose diagonal elements $\mathbf{D}_{i,i}=\sum_j \mathbf{A}_{i,j}$ are the degrees of corresponding nodes.

Given the preprocessed adjacency matrix $\mathbf{\hat{A}}$ and the initial graph signal $\mathbf{X}$ as inputs, we apply the GCN propagation process as follows:
\begin{equation}
\mathbf{H} = ReLU(\mathbf{\hat{A}}~ReLU(\mathbf{\hat{A}} \mathbf{X} \mathbf{W}_a^{(1)}) \mathbf{W}_a^{(2)})
\label{equ:gcn}
\end{equation}
where $ReLU(\cdot)$ denotes the ReLU activation function, $\mathbf{W}_a^{(1)}$ and $\mathbf{W}_a^{(2)}$ are two weight parameters at the $1$-th and $2$-th layer of the GCN encoder, respectively. Specifically, this encoder directly captures affinity relationships between drugs and targets at the first GCN propagation iteration, which can be regarded as a spatial-based graph filter aggregating directly connected neighbours (i.e., $1$-hop neighbours) to update node representations. The second GCN propagation aggregates the information of $2$-hop neighbour nodes to recognize the hidden similarity patterns, whose chemical interpretability is based on the empirical assumption that chemically, biologically, or topologically similar drugs are more likely to interact with the same target and vice versa. After two iterations, the GCN encoder generates the global-level embedding matrix $\mathbf{H} \in \mathbb{R}^{(n_d + n_t) \times d_{ag}^\prime}$, whose row vectors correspond to drugs $\mathcal{V}_d = \{d_1, \dots, d_{n_d}\}$ and targets $\mathcal{V}_t = \{t_1, \dots, t_{n_t}\}$ in the affinity graph $\mathcal{N}$, $d_{ag}^\prime$ is the embedding dimension.

Assumed that a drug-target pair $(d_i, t_j)$ is selected from the training set $\mathcal{T}$, we extract the global-level representations $\mathbf{h}_{d_i} \in \mathbb{R}^{d_{dg}}$ for drug $d_i$ and $\mathbf{h}_{t_j} \in \mathbb{R}^{d_{tg}}$ for target $t_j$, respectively, using two non-linear Multi Layer Perceptron (MLP) operators:
\begin{equation}
\mathbf{h}_{d_i} = MLP(\mathbf{W}_b^{(1)}, \mathbf{W}_b^{(2)}, \mathbf{H}[i,:]),~\mathbf{h}_{t_j} = MLP(\mathbf{W}_c^{(1)}, \mathbf{W}_c^{(2)}, \mathbf{H}[j+n_d,:])
\label{equ:transform}
\end{equation}
where $\mathbf{H}[i,:]$ and $\mathbf{H}[j+n_d,:]$ denote the row vectors in the embedding matrix $\mathbf{H}$ corresponding to $d_i$ and $t_j$ respectively, $\mathbf{W}_b^{(1)}$, $\mathbf{W}_b^{(2)}$, $\mathbf{W}_c^{(1)}$ and $\mathbf{W}_c^{(2)}$ are the weight parameters of the MLP operators.
}

\subsection{Local-level graph representation learning on molecular graph}
For each molecular graph, we equip our model with self-loop GCNs to learn local-level molecular representations. In the GCN propagation process, we smooth the biological features of each node (i.e., atom or residue) over the molecular graph to map the intrinsic spatial structures into the low-dimensional embedding space.{

Mathematically, given a molecular graph $\mathcal{G} = \{\mathcal{V}, \mathcal{E}\}$, where $\mathcal{V}$ is the set of nodes and $\mathcal{E}$ is the set of edges, we introduce the self-loop GCN propagation operator $f_{prop}(\cdot)$ over the molecular graph:
\begin{equation}
\mathbf{h}_{v_i}^{(l)} = f_{prop}(\{\mathbf{h}_{v_j}^{(l-1)}~\mid~v_j \in C(v_i) \cup \{v_i\}\}) = ReLU(\sum_{v_j} \frac{1}{\sqrt{\hat{d}_{v_i} \hat{d}_{v_j}}} \mathbf{h}_{v_j}^{(l-1)} \mathbf{W}_d^{(l)})
\label{equ:node_level_gcn}
\end{equation}
where $v_i$ represents the $i$-th node in the graph $\mathcal{G}$, $C(v_i)$ is the neighbours of the node $v_i$, $\mathbf{h}_{v_i}^{(l)}$ denotes the hidden state at the $l$-th graph convolutional layer for node $v_i$, $\mathbf{h}_{v_i}^{(0)}$ denotes the initial feature of node $v_i$, $\hat{d}_{v_i} = 1 + \mid C(v_i) \mid$ denotes the degree of node $v_i$, $\mathbf{W}_d^{(l)}$ is the weight parameter at the $l$-th layer. It should be noted that the self-loop is added into the calculation of the degree $\hat{d}_{v_i}$, where the node $v_i$ itself is viewed as its 1-hop neighbour to be taken into consideration. Through the self-loop operation, we can take full advantage of the biological features of nodes (i.e., atoms or residues) during the GCN propagation.

In order to learn local-level molecular representations of drugs/targets, we apply the above self-loop GCN propagation framework over the drug molecular graph and the target molecular graph, respectively. To be specific, given the drug $d_i$ and its corresponding drug molecular graph $\mathcal{G}_{d_i} = \{\mathcal{V}_a, \mathcal{E}_b\}$, we apply the self-loop convolutional operator over the graph's topology with the initial attributions $\{\mathbf{x}_{a_u}\}$ of atoms $\mathcal{V}_a = \{a_1, \dots, a_{n_a}\}$ as input. On this basis, we encode multiple atomic features and chemical structure information of drug molecules into the local-level atom representations $\{\mathbf{h}_{a_u} \in \mathbb{R}^{d_{dg}}\}$ through Equation (\ref{equ:node_level_gcn}). Similarly, for the target molecular graph $\mathcal{G}_{t_j} = \{\mathcal{V}_r, \mathcal{E}_c\}$ corresponding to the target $t_j$, we take the initial features $\{\mathbf{x}_{r_p}\}$ of residue $\mathcal{V}_r = \{r_1, \dots, r_{n_r}\}$ as input to obtain the local-level residue representations $\{\mathbf{h}_{r_p} \in \mathbb{R}^{d_{tg}}\}$ through Equation (\ref{equ:node_level_gcn}). Note that the parameters of GCNs on the molecular graph are shared across all the atoms or residues.
}

\subsection{Integrating hierarchical graph information via message broadcasting}
After the hierarchical graph representation learning introduced above, for drug/target molecules, we obtain their global-level representations deriving from the affinity graph and local-level representations learned from the molecular graph, respectively. In order to integrate the hierarchical graph information, we make use of a message broadcasting mechanism to encode the global-level representations of drugs/targets into their corresponding local-level molecular representations, and reuse GCN propagation to refine the molecular representations. It is worth noticing that the global-level affinity information is used to guide the representation learning of the local-level molecular graph via the refinement process.{

In the message broadcasting strategy, a sender can deliver information to plenty of recipients simultaneously. In our model, the global-level drug/target embedding and its corresponding local-level atom/residue embeddings are regarded a sender and recipients. Through message broadcasting, the global-level representation of a drug (or target) is shared by all of the atoms (or residues) to update their local-level representations. Concretely, given drug embedding $\mathbf{h}_{d_i}$, target embedding $\mathbf{h}_{t_j}$, atom representations $\{\mathbf{h}_{a_u}\}$ and residue representations $\{\mathbf{h}_{r_p}\}$, the message broadcasting process can be formulated as follows:
\begin{equation}
\{\mathbf{z}_{a_u}\} = \{\mathbf{h}_{a_u}~\oplus~\mathbf{h}_{d_i}~\mid\mid~\mathbf{h}_{a_u}~\ominus~\mathbf{h}_{d_i}\},~
\{\mathbf{z}_{r_p}\} = \{\mathbf{h}_{r_p}~\oplus~\mathbf{h}_{t_j}~\mid\mid~\mathbf{h}_{r_p}~\ominus~\mathbf{h}_{t_j}\}
\label{equ:message_broadcasting}
\end{equation}
where $\{\mathbf{z}_{a_u} \in \mathbb{R}^{2 \cdot d_{dg}}\}$ and $\{\mathbf{z}_{r_p} \in \mathbb{R}^{2 \cdot d_{tg}}\}$ denote the updated atom embeddings and residue embeddings, $\oplus$ and $\ominus$ denote element-wise addition and subtraction respectively, $\mid\mid$ denotes the concatenation operation. In particular, the element-wise addition represents the fusion of global-level and local-level information and the element-wise subtraction explores the difference between them.

Through the message broadcasting process, we can integrate the global-level affinity information into the local-level molecular graphs to update atom and residue embeddings. Whereas, this update process may cause the chemical structure information blurred in the molecular representations. Thus, it is necessary to reload the molecular chemical structures in our model. For this reason, we reuse GCNs over the molecular graph's topology to refine the updated atom and residue embeddings. To be specific, given the updated atom embeddings $\{\mathbf{z}_{a_u}\}$ and residue embeddings $\{\mathbf{z}_{r_p}\}$ as input, we generate the refined representations $\{\mathbf{\hat{z}}_{a_u} \in \mathbb{R}^{d_{dg}^\prime}\}$ for atoms and $\{\mathbf{\hat{z}}_{r_p} \in \mathbb{R}^{d_{tg}^\prime}\}$ for residues through Equation (\ref{equ:node_level_gcn}), where $\mathbb{R}^{d_{dg}^\prime}$ and $\mathbb{R}^{d_{tg}^\prime}$ are the representation dimensions.

We readout the final drug embedding $\mathbf{d}_i$ from the drug molecular graph $G_{d_i}$ through an average pooling operation along the atom dimension. Similarly, we readout the final target embedding $\mathbf{t}_j$ through combining representation vectors across all residues in the target molecular graph $G_{t_j}$. The readout layer is formulated as follows:
\begin{equation}
\mathbf{d}_i = MLP(\mathbf{W}_e^{(1)}, \mathbf{W}_e^{(2)}, f_{avg}(\{\mathbf{\hat{z}}_{a_u}\})),~\mathbf{t}_j = MLP(\mathbf{W}_f^{(1)}, \mathbf{W}_f^{(2)}, f_{avg}(\{\mathbf{\hat{z}}_{r_p}\}))
\label{equ:readout}
\end{equation}
where $f_{avg}$ denotes the average pooling operation, $\mathbf{W}_e^{(1)}$, $\mathbf{W}_e^{(2)}$, $\mathbf{W}_f^{(1)}$ and $\mathbf{W}_f^{(2)}$ are the weight parameters of the MLP operators.
}

\subsection{Drug-target binding affinity prediction}
In this study, the binding affinity prediction problem is treated as a regression task. For each drug-target pair $(d_i, t_j) \in \mathcal{T}$, we combine the drug embedding $\mathbf{d}_i$ and the target embedding $\mathbf{t}_j$ into an affinity embedding vector and then make the final affinity prediction using a three-layer MLP:
\begin{equation}
y_{i,j} = MLP(\mathbf{W}_g^{(1)}, \mathbf{W}_g^{(2)}, \mathbf{W}_g^{(3)}, \mathbf{d}_i~\mid\mid~\mathbf{t}_j)
\label{equ:predictor}
\end{equation}
where $y_{i,j}$ is the predicted affinity value of the drug-target pair $(d_i, t_j)$, $\mid\mid$ denotes the concatenation operation, $\mathbf{W}_g^{(1)}$, $\mathbf{W}_g^{(2)}$ and $\mathbf{W}_g^{(3)}$ are the weight parameters of the MLP operator.

Given a batch of drug-target pairs in the training set $\mathcal{T}$ and their corresponding ground-truth affinity values, we train the weight parameters with the following MSE loss function:
\begin{equation}
\mathcal{L} = \frac1m \sum (y_{i,j} - \hat{y}_{i,j}) ^ 2
\label{equ:loss_function}
\end{equation}
where $m$ represents the number of drug-target pairs in a training batch, $\hat{y}_{i,j}$ denotes the ground-truth affinity value of the drug-target pair $(d_i, t_j)$. Based on the loss function, we optimize the mapping function $\mathit{\Theta}(\omega)$ using the back-propagation algorithm and the Adam \cite{Kingma2015} optimizer, to find the best solution for the trainable parameter $\omega$.

\renewcommand{\algorithmicrequire}{\textbf{Input:}}  
\renewcommand{\algorithmicensure}{\textbf{Output:}}  
\begin{algorithm}[htpb]
    \caption{The Hierarchical Graph Representation Learning Algorithm}
    \label{alg:HGRL-DTA}
    \begin{algorithmic}[1]
        \Require Training set $\mathcal{T}$; affinity graph $\mathcal{N}$; drug molecular graphs $\{\mathcal{G}_{d_i}\}_{i=1}^{n_d}$; target molecular graphs $\{\mathcal{G}_{t_j}\}_{j=1}^{n_t}$
        \Ensure Mapping function $\mathit{\Theta}(\omega)$
        \While{HGRL-DTA not converge}
            \State $\mathbf{\hat{A}}$ $\gets$ preprocess the adjacency matrix via Eq.(\ref{equ:adj_preprocess})
            \State $\mathbf{H}$ $\gets$ calculate the global-level embedding matrix via Eq.(\ref{equ:gcn})
            \For{$(d_i, t_j) \in \mathcal{T}$}
                \State $\mathbf{h}_{d_i}, \mathbf{h}_{t_j}$ $\gets$ calculate the global-level representations of the drug and the target via Eq.(\ref{equ:transform})
                \For{$a_u \in \mathcal{V}_a$}
                    \State $\mathbf{h}_{a_u}$ $\gets$ calculate the local-level atom representation via Eq.(\ref{equ:node_level_gcn})
                \EndFor
                \For{$r_p \in \mathcal{V}_r$}
                    \State $\mathbf{h}_{r_p}$ $\gets$ calculate the local-level residue representation via Eq.(\ref{equ:node_level_gcn})
                \EndFor
                \State $\{\mathbf{z}_{a_u}\}, \{\mathbf{z}_{r_p}\}$ $\gets$ update the representations of atoms and residues via Eq.(\ref{equ:message_broadcasting})
                \State $\{\mathbf{\hat{z}}_{a_u}\}$ $\gets$ refine the atom representations like \textbf{line 6-8}
                \State $\{\mathbf{\hat{z}}_{r_p}\}$ $\gets$ refine the residue representations like \textbf{line 9-11}
                \State $\mathbf{d}_i, \mathbf{t}_j$ $\gets$ readout the final drug embedding and target embedding via Eq.(\ref{equ:readout})
                \State $y_{i,j}$ $\gets$ predict the binding affinity via Eq.(\ref{equ:predictor})
            \EndFor
            \State $\mathcal{L}$ $\gets$ calculate the loss function via Eq.(\ref{equ:loss_function})
            \State $\nabla_{\omega} \mathcal{L}$
        \EndWhile \\
        \Return $\mathit{\Theta}(\omega)$
    \end{algorithmic}
\end{algorithm}

\section{Experiment}
In this section, we firstly present the datasets, experimental settings, and evaluation metrics used in our experiments. Then, we describe several important implementation details of our proposed model. Afterwards, we compare the HGRL-DTA model with other state-of-the-art methods under different experimental scenarios. Finally, we make further and deeper analyses on HGRL-DTA with ablation study, parameter analysis and visualization analysis.

\subsection{Datasets}
\label{sec:datasets}
To evaluate the performance of our proposed model on the binding affinity prediction task, two classic benchmark datasets, the Davis dataset \cite{Davis2011} and the KIBA dataset \cite{Tang2014}, are chosen to conduct experiments.

\textbf{Davis.} The Davis dataset contains 68 unique drugs and 442 unique targets, with 30,056 kinase dissociation constant $K_d$ values as drug-target affinities. He et al. \cite{He2017} transform $K_d$ values in the Davis dataset into log space as: $pK_d = - \log_{10} (K_d / {10}^9)$. The preprocessed Davis dataset is filled with affinities ranging from 5.0 to 10.8, where the boundary value 5.0 is regarded as the true negative drug-target pair that either has very weak binding affinities or is not detected in the wet lab experiment. The Davis dataset collected drugs' SMILES strings from the PubChem compound database \cite{Bolton2008} based on PubChem CIDs, and targets' protein sequences from the UniProt protein database \cite{Apweiler2004} according to gene names/RefSeq accession numbers.

\textbf{KIBA.} The KIBA dataset introduces KIBA scores as drug-target affinities, based on the integration of kinase inhibitor bioactivities from various sources such as $K_i$, $K_d$, and $IC_{50}$ \cite{Tang2014}. The dataset originally consists of 52,498 drugs and 467 targets with 246,088 affinities. He et al. \cite{He2017} filtered it to comprise 118,254 affinities between 2,111 unique drugs and 229 unique targets with 10 affinities at least of each drug and target. The preprocessed KIBA dataset contains affinity values ranging from 0.0 to 17.2, and nan values indicating there are no experimental values for corresponding drug-target pairs. The KIBA dataset converted drugs' CHEMBL IDs \cite{Gaulton2012} into their corresponding PubChem CIDs to extract SMILES strings based on PubChem CIDs and collected protein sequences for targets through UniProt IDs.

\subsection{Experimental settings and evaluation metrics}
Following previous works \cite{Jiang2020, Ozturk2018}, we evaluate the performance of various models with the division ratio for training and testing as 5:1. When conducting experiments, we train the proposed model and compared methods on the training set and evaluate them on the test set. We conduct 5-fold cross validation (CV) on the training set to select the best hyper-parameters from fixed ranges for our proposed model. For all the compared methods, their hyper-parameters are set as the optimal values provided in the corresponding studies \cite{Jiang2020, Nguyen2020, Ozturk2018, Zhao2019}.

In this study, to verify the generalization and robustness of the models comprehensively, we consider the following four experimental scenarios \cite{Pahikkala2014}:

\begin{itemize}[leftmargin=11mm, itemsep=0.05mm]
\item[$\bullet$] \textbf{S1:} Entries in the drug-target matrix $\mathbf{Y}$ are randomly selected for testing.
\item[$\bullet$] \textbf{S2:} Row vectors in the drug-target matrix $\mathbf{Y}$ are randomly selected for testing.
\item[$\bullet$] \textbf{S3:} Column vectors in the drug-target matrix $\mathbf{Y}$ are randomly selected for testing.
\item[$\bullet$] \textbf{S4:} The intersection set of the row vectors in the setting \textbf{S2} and the column vectors in the setting \textbf{S3} is selected for testing, and their non-intersection parts are used for neither training nor testing.
\end{itemize}

In setting \textbf{S1}, both the drug and the target of the test drug-target pairs can be observed in the training phase. As the most widely used experimental setting in previous studies, the setting \textbf{S1} assumes that a part of known drug-target affinities are randomly masked and our aim is to infer these masked affinities using other known ones. Compared to setting \textbf{S1}, the settings \textbf{S2} and \textbf{S3} have attracted more attention in real-world applications recently, where only part of drug/target information is available during training and the models need to predict affinities for new drugs/targets which do not bind any known affinities. The setting \textbf{S4} corresponds to the most challenging case in computational works, which aims to predict affinities between unknown drugs and targets.

We choose four classic metrics to evaluate the performance of the models: Mean Squared Error (MSE), Concordance Index (CI) \cite{Gonen2005}, $r_m^2$ \cite{PratimRoy2009, Roy2013} and Pearson correlation coefficient (Pearson) \cite{Benesty2009}. For each model, we report the average and the standard deviation (std) of these indicators across ten random runs.

\subsection{Implementation details}
\label{sec:implementation_details}
In this subsection, we present the important implementation details of the proposed model, which correspond to GNNs' practical issues including constructing graphs, inferring representations for unseen nodes, and preventing over-fitting and over-smoothing. Then we describe the runtime environment for modeling and conducting experiments.

\subsubsection{Constructing graphs}
In the following, we introduce the construction process of the input graphs, i.e., the affinity graph, the drug molecular graph, and the target molecular graph, defined in Section \ref{sec:preliminaries}.

\textbf{Affinity graph.} For the affinity graph, we normalize raw affinity values into the range [0, 1] using min-max normalization and formulate normalized affinities as its corresponding adjacency matrix. Each node (i.e., drug or target) in the affinity graph is encoded as a multi-dimensional binary feature vector, which consists of two pieces of information: one-hot encoding of the node type (i.e., either drug-type or target-type) and one-hot encoding of the neighbour nodes (i.e., row vector in the connectivity matrix corresponding to the affinity graph).

\textbf{Drug molecular graph.} We transform drugs' SMILES (Simplified Molecular Input Line Entry System) strings \cite{Weininger1988}, which were invented to represent molecules to be readable by computers, into their corresponding graphs with the open-source chemical informatics software RDKit \cite{Landrum2016}. A group of atomic features adopted from DeepChem \cite{Ramsundar2019} is used as the initial drug molecular graph signals.

\textbf{Target molecular graph.} Targets' complex folded structure are composed of a variety of spatial characteristics, which contain residues, peptide bonds, and non-bonded interactions such as hydrogen bonds and van der Waals forces. However, obtaining the tertiary structures of protein by crystallization in the laboratory is time-consuming and costly, which results in that there exist a large number of protein structures unavailable. To handle this issue, Pconsc4 \cite{Michel2018}, an open-source and highly efficient protein structure prediction approach, is adopted in our work to generate target molecular graphs through mining useful topological information hidden in protein sequences. The Pconsc4 method transforms targets' protein sequences into their corresponding contact maps, i.e., residue-residue association matrixes, whose entries are the Euclidean distance-based contacts. In this contact map, there exists a contact between two atoms if the Euclidean distance between them is less than a specified threshold \cite{Wu2019}. We set the threshold as 0.5 according to the previous study DGraphDTA \cite{Jiang2020}. A set of residue features extracted by DGraphDTA \cite{Jiang2020} is used as the initial attributions of residues in the target molecular graph.

\subsubsection{Inferring representations for unseen nodes}
In view of the affinity graph, the GCN encoder works based on the graph connectivity of the affinity data. However, there exists a challenge that this encoder cannot learn representations for unseen nodes (i.e., unknown drugs/targets). Such a cold start problem occurs when conducting experiments in the settings \textbf{S2}, \textbf{S3}, and \textbf{S4}, where a part of drugs or targets are not observed in the training phase. This problem limits the generalization ability of our model for unseen drugs and targets.

To solve this problem, we introduce drug-drug similarities and target-target similarities to construct a \textbf{similarity-based embedding map} from known to unknown drugs/targets, which can infer global-level representations for unknown drugs/targets during testing. In detail, we firstly calculate structural fingerprint similarities between known and unknown drugs using PubChem Score Matrix Service \cite{Bolton2008} and sequence similarities between known and unknown targets using Smith-Waterman (SW) algorithm \cite{Smith1981}, respectively. Then, take the drug as an example (target is similar), we consider a bipartite similarity graph, in which each unknown drug is connected to its $simK$ most similar known drugs. We implement the embedding map to infer an unknown drug by aggregating the learned global-level representations of its neighbours with a weighted summation operation, which utilizes normalized similarities as weights. Through this map, we can infer the representations of unseen drugs and targets to generalize our model for the cold start situation. In this paper, we set the $simK$ of drug $simK_d = 2$ and target $simK_t = 7$, respectively.

Compared to the learned representations of known drugs and targets, the inferred representations of unseen drugs and targets contain less topological affinity information. Moreover, the model may lose some global-level affinity information when integrating the hierarchical graph representations. Therefore, in order to allow sufficient global-level affinity information to be retained, we add a skip connection from the global-level drug/target embedding to the readout layer. It should be noted that we only conduct the above operations under settings \textbf{S2}, \textbf{S3}, and \textbf{S4}.

\subsubsection{Preventing over-fitting and over-smoothing}
Over-fitting and over-smoothing frequently occur during training GNN-based models. To alleviate these issues, we introduce a regularization technique DropEdge \cite{Rong2020}, which randomly removes a certain number of edges from the input graph at each training epoch, into the graph convolutional process over the affinity graph. Also, it has been theoretically demonstrated that DropEdge either reduces the convergence speed of over-smoothing or relieves the information loss caused by it \cite{Rong2020}. In this paper, we set the DropEdge rate $\alpha = 0.2$.

In addition, it has been theoretically and empirically proved that nodes with high degrees are more likely to suffer from over-smoothing in multi-layer GNN-based models \cite{Chen2020b}. In the KIBA dataset, we observe that each target connects with too many (on average 518 and up to 1,452) drugs, which causes the over-smoothing problem occurring. To handle this issue, we selectively dropout edges of the affinity graph in the data preprocessing phase, which only preserves the $topK$ highest affinity edges related to each target and removes other connected ones. In this paper, we set the $topK$ of target $topK_t = 150$ under settings \textbf{S1}, \textbf{S3} and $topK_t = 90$ under settings \textbf{S2}, \textbf{S4}. Without loss of generality, we perform the same operation for each drug and set the $topK$ of drug $topK_d = 40$. Note that we selectively remove affinities only when conducting experiments on the KIBA dataset.\vspace{5pt}

We implement our proposed model with Pytorch 1.4.0 \cite{Paszke2019} and Pytorch-geometric 1.7.0 \cite{Fey2019}. We run HGRL-DTA on our workstation with 2 Intel(R) Xeon(R) Gold 6146 3.20GHZ CPUs, 128GB RAM, and 2 NVIDIA 1080 Ti GPUs. For more detailed parameter settings of HGRL-DTA, please refer to the source code: \url{https://github.com/Zhaoyang-Chu/HGRL-DTA}.

\subsection{Comparison with state-of-the-art methods}
To demonstrate the superiority of the proposed model, we conduct experiments to compare our approach with the following state-of-the-art methods:
\begin{itemize}[leftmargin=11mm, itemsep=0.05mm]
\item[$\bullet$] \textbf{DeepDTA} \cite{Ozturk2018} employs CNNs to learn 1D drug and target representations from drug SMILES strings and target protein sequences.
\item[$\bullet$] \textbf{AttentionDTA} \cite{Zhao2019} utilizes 1D CNNs to learn sequence representations of drugs and targets and an attention mechanism to find the weight relationships between drug subsequences and protein subsequences.
\item[$\bullet$] \textbf{GraphDTA} \cite{Nguyen2020} models drugs as molecular graphs to capture the bonds among atoms with GNNs and leverages CNNs to learn 1D representations of target proteins.
\item[$\bullet$] \textbf{DGraphDTA} \cite{Jiang2020} constructs target molecular graphs from the corresponding protein sequences via the protein structure prediction method and applies GNNs to mine structural information hidden in drug molecular graphs and target molecular graphs.
\end{itemize}

\begin{table}[htbp]
\scriptsize
\setlength\tabcolsep{3pt}
\caption{Performances of HGRL-DTA and compared methods.}
\centering
\begin{tabular}{ll|llll|llll}
    \toprule
    \multicolumn{2}{c|}{\multirow{2}{*}{Architecture}} & \multicolumn{4}{c|}{Davis} & \multicolumn{4}{c}{KIBA} \\ 
    \cmidrule(lr){3-10}
    & & \makecell[c]{MSE$\downarrow$ (std)} & \makecell[c]{CI$\uparrow$ (std)} & \makecell[c]{$r_m^2$$\uparrow$ (std)} & \makecell[c]{Pearson$\uparrow$ (std)} & \makecell[c]{MSE$\downarrow$ (std)} & \makecell[c]{CI$\uparrow$ (std)} & \makecell[c]{$r_m^2$$\uparrow$ (std)} & \makecell[c]{Pearson$\uparrow$ (std)} \\
    \midrule
    \multirow{5}{*}{\textbf{S1}}
        & DeepDTA & \makecell[c]{0.245 (0.014)} & \makecell[c]{0.888 (0.004)} & \makecell[c]{0.665 (0.015)} & \makecell[c]{0.842 (0.004)} & \makecell[c]{0.181 (0.007)} & \makecell[c]{0.868 (0.004)} & \makecell[c]{0.711 (0.021)} & \makecell[c]{0.864 (0.003)} \\
        & AttentionDTA & \makecell[c]{0.233 (0.006)} & \makecell[c]{0.889 (0.002)} & \makecell[c]{0.676 (0.020)} & \makecell[c]{0.845 (0.004)} & \makecell[c]{0.150 (0.002)} & \makecell[c]{0.883 (0.001)} & \makecell[c]{0.760 (0.018)} & \makecell[c]{0.888 (0.001)} \\
        & GraphDTA & \makecell[c]{0.243 (0.005)} & \makecell[c]{0.887 (0.002)} & \makecell[c]{0.685 (0.016)} & \makecell[c]{0.839 (0.003)} & \makecell[c]{0.148 (0.006)} & \makecell[c]{0.891 (0.001)} & \makecell[c]{0.730 (0.015)} & \makecell[c]{0.895 (0.001)} \\
        & DGraphDTA & \makecell[c]{0.216 (0.003)} & \makecell[c]{0.900 (0.001)} & \makecell[c]{0.686 (0.015)} & \makecell[c]{0.857 (0.002)} & \makecell[c]{0.132 (0.002)} & \makecell[c]{0.902 (0.001)} & \makecell[c]{\textbf{0.800 (0.011)}} & \makecell[c]{0.903 (0.001)} \\
        & HGRL-DTA & \makecell[c]{\textbf{0.166 (0.002)}} & \makecell[c]{\textbf{0.911 (0.002)}} & \makecell[c]{\textbf{0.751 (0.006)}} & \makecell[c]{\textbf{0.892 (0.001)}} & \makecell[c]{\textbf{0.125 (0.001)}} & \makecell[c]{\textbf{0.906 (0.001)}} & \makecell[c]{0.789 (0.017)} & \makecell[c]{\textbf{0.907 (0.001)}} \\
    \midrule
    \multirow{5}{*}{\textbf{S2}}
        & DeepDTA & \makecell[c]{0.985 (0.114)} & \makecell[c]{0.548 (0.045)} & \makecell[c]{0.027 (0.022)} & \makecell[c]{0.126 (0.109)} & \makecell[c]{0.494 (0.070)} & \makecell[c]{0.747 (0.012)} & \makecell[c]{0.337 (0.026)} & \makecell[c]{0.623 (0.023)} \\
        & AttentionDTA & \makecell[c]{0.869 (0.053)} & \makecell[c]{0.642 (0.028)} & \makecell[c]{0.079 (0.024)} & \makecell[c]{0.289 (0.048)} & \makecell[c]{0.506 (0.018)} & \makecell[c]{0.744 (0.005)} & \makecell[c]{0.298 (0.015)} & \makecell[c]{0.618 (0.006)} \\
        & GraphDTA & \makecell[c]{0.801 (0.038)} & \makecell[c]{0.659 (0.015)} & \makecell[c]{0.160 (0.019)} & \makecell[c]{0.416 (0.022)} & \makecell[c]{0.475 (0.047)} & \makecell[c]{0.753 (0.002)} & \makecell[c]{\textbf{0.382 (0.007)}} & \makecell[c]{0.652 (0.002)} \\
        & DGraphDTA & \makecell[c]{0.818 (0.012)} & \makecell[c]{0.646 (0.006)} & \makecell[c]{0.114 (0.005)} & \makecell[c]{0.356 (0.010)} & \makecell[c]{0.458 (0.008)} & \makecell[c]{0.754 (0.002)} & \makecell[c]{0.362 (0.012)} & \makecell[c]{0.622 (0.004)} \\
        & HGRL-DTA & \makecell[c]{\textbf{0.776 (0.019)}} & \makecell[c]{\textbf{0.684 (0.007)}} & \makecell[c]{\textbf{0.163 (0.015)}} & \makecell[c]{\textbf{0.422 (0.018)}} & \makecell[c]{\textbf{0.434 (0.007)}} & \makecell[c]{\textbf{0.757 (0.003)}} & \makecell[c]{0.370 (0.010)} & \makecell[c]{\textbf{0.653 (0.003)}} \\
    \midrule
    \multirow{5}{*}{\textbf{S3}}
        & DeepDTA & \makecell[c]{0.552 (0.086)} & \makecell[c]{0.729 (0.017)} & \makecell[c]{0.258 (0.029)} & \makecell[c]{0.523 (0.028)} & \makecell[c]{0.732 (0.197)} & \makecell[c]{0.676 (0.016)} & \makecell[c]{0.273 (0.026)} & \makecell[c]{0.587 (0.033)} \\
        & AttentionDTA & \makecell[c]{0.436 (0.017)} & \makecell[c]{0.787 (0.018)} & \makecell[c]{0.304 (0.022)} & \makecell[c]{0.588 (0.027)} & \makecell[c]{0.529 (0.039)} & \makecell[c]{0.693 (0.008)} & \makecell[c]{0.254 (0.024)} & \makecell[c]{0.592 (0.022)} \\
        & GraphDTA & \makecell[c]{0.860 (0.083)} & \makecell[c]{0.666 (0.012)} & \makecell[c]{0.134 (0.014)} & \makecell[c]{0.379 (0.018)} & \makecell[c]{0.469 (0.089)} & \makecell[c]{0.710 (0.005)} & \makecell[c]{0.388 (0.013)} & \makecell[c]{0.627 (0.009)} \\
        & DGraphDTA & \makecell[c]{0.445 (0.019)} & \makecell[c]{0.788 (0.009)} & \makecell[c]{0.289 (0.016)} & \makecell[c]{0.558 (0.017)} & \makecell[c]{0.364 (0.010)} & \makecell[c]{0.718 (0.007)} & \makecell[c]{0.429 (0.022)} & \makecell[c]{0.671 (0.009)} \\
        & HGRL-DTA & \makecell[c]{\textbf{0.383 (0.010)}} & \makecell[c]{\textbf{0.816 (0.008)}} & \makecell[c]{\textbf{0.375 (0.018)}} & \makecell[c]{\textbf{0.621 (0.012)}} & \makecell[c]{\textbf{0.322 (0.014)}} & \makecell[c]{\textbf{0.741 (0.004)}} & \makecell[c]{\textbf{0.502 (0.016)}} & \makecell[c]{\textbf{0.729 (0.007)}} \\
    \midrule
    \multirow{5}{*}{\textbf{S4}}
        & DeepDTA & \makecell[c]{0.767 (0.091)} &\makecell[c]{0.508 (0.057)} & \makecell[c]{0.009 (0.012)} & \makecell[c]{0.015 (0.098)} & \makecell[c]{0.700 (0.075)} &\makecell[c]{0.627 (0.009)} & \makecell[c]{0.140 (0.017)} & \makecell[c]{0.401 (0.025)} \\
        & AttentionDTA & \makecell[c]{0.679 (0.021)} &\makecell[c]{0.554 (0.030)} & \makecell[c]{0.005 (0.008)} & \makecell[c]{0.036 (0.062)} & \makecell[c]{0.609 (0.021)} &\makecell[c]{0.629 (0.007)} & \makecell[c]{0.143 (0.015)} & \makecell[c]{0.407 (0.022)} \\
        & GraphDTA & \makecell[c]{0.988 (0.096)} &\makecell[c]{0.569 (0.017)} & \makecell[c]{0.020 (0.006)} & \makecell[c]{0.141 (0.020)} & \makecell[c]{0.676 (0.113)} &\makecell[c]{0.641 (0.003)} & \makecell[c]{0.149 (0.007)} & \makecell[c]{0.404 (0.009)} \\
        & DGraphDTA & \makecell[c]{0.658 (0.026)} &\makecell[c]{0.569 (0.008)} & \makecell[c]{0.031 (0.005)} & \makecell[c]{0.180 (0.015)} & \makecell[c]{0.594 (0.022)} & \makecell[c]{0.632 (0.009)} & \makecell[c]{0.148 (0.013)} & \makecell[c]{0.403 (0.019)} \\
        & HGRL-DTA & \makecell[c]{\textbf{0.642 (0.016)}} & \makecell[c]{\textbf{0.602 (0.009)}} & \makecell[c]{\textbf{0.044 (0.005)}} & \makecell[c]{\makecell[c]{\textbf{0.215 (0.013)}}} & \makecell[c]{\textbf{0.532 (0.008)}} & \makecell[c]{\textbf{0.642 (0.004)}} & \makecell[c]{\textbf{0.207 (0.009)}} & \makecell[c]{\makecell[c]{\textbf{0.491 (0.010)}}} \\
    \bottomrule
\end{tabular}
\label{tab:table_comparison}
\end{table}

Table \ref{tab:table_comparison} compares our HGRL-DTA model's performance against the state-of-the-art methods on the two benchmark datasets under four experimental scenarios. We highlight the best results in boldface. According to the experimental results, we can observe that the proposed HGRL-DTA model achieves the best performance compared to the state-of-the-art methods under all the scenarios, which demonstrates the generalization and robustness of our model. In four experimental settings, over the best baseline models, we achieve 23.1\%, 3.1\%, 12.2\% and 2.4\% improvement of MSE on the Davis dataset, 5.3\%, 5.2\%, 11.5\% and 10.4\% improvement of MSE on the KIBA dataset. In most cases, the standard deviations of the HGRL-DTA model are lower than other compared methods, which demonstrates the stability of the predictive model.

Among all baselines, the performance of the sequence-based methods (i.e., DeepDTA, AttentionDTA) is relatively poor due to the inadequate exploitation of the molecular chemical structures. It indicates that simply modeling drugs as SMILES strings and targets as protein sequences is not a natural way to capture the intrinsic properties of molecules. By contrast, the graph-based models (i.e., GraphDTA and DGraphDTA) represent molecules as molecular graphs to take advantage of their chemical structural information, which produces better predictive performance. However, these graph-based models mainly focus on encoding molecular structures but ignore the abundant topological information deriving from drug-target affinity relationships, which causes them inferior to our proposed model. Compared with the state-of-the-art models, HGRL-DTA can capture the local-level intrinsic molecular properties and the global-level topological affinity relationships simultaneously and incorporate such hierarchical information into the representations of drugs and targets, which significantly facilitates the performance of predicting drug-target binding affinities.

In addition, we observe that all the models perform best in setting \textbf{S1}, have relatively poor performance in settings \textbf{S2} and \textbf{S3}, and perform worst in setting \textbf{S4}. With more unknown drugs or targets in the four experimental settings, the predictive performance of the models significantly declines. Different from setting \textbf{S1}, the settings \textbf{S2}, \textbf{S3} and \textbf{S4} test the generalization and robustness of the models for unseen drugs or targets, which is another necessary measurement of performance evaluation. Through the similarity-based embedding map, HGRL-DTA infers unseen drugs/targets using the learned embeddings of known drugs/targets, which can make full use of the known affinity and similarity information to improve the generalization and robustness of the model. As illustrate in Table \ref{tab:table_comparison}, the HGRL-DTA model obtains the best performance in the settings \textbf{S2}, \textbf{S3} and \textbf{S4}, which indicates that HGRL-DTA is more generalizable and more robust compared to baseline methods when only part of drug/target information is known.

\subsection{Ablation study}
\label{sec:ablation_study}
To investigate the important factors that impact the predictive capacity of our model, we conduct the ablation study with the following variants of HGRL-DTA in setting \textbf{S1}:
\begin{itemize}[leftmargin=11mm, itemsep=0.05mm]
\item[$\bullet$] \textbf{HGRL-DTA without global-level affinity graph} (w/o GAG) only learns local-level representations on the molecular graph without the GCN propagation on the global-level affinity graph. Note that this variant keeps the same number of GCN iterations on the molecular graph as HGRL-DTA, where GCN iterations in the refinement procedure of HGRL-DTA are also considered.
\item[$\bullet$] \textbf{HGRL-DTA without local-level molecular graph} (w/o LMG) only applies the GCN propagation on the global-level affinity graph without considering the local-level molecular graphs. The MLP-based predictor is directly applied with the input of the global-level representations of drugs and targets for the binding affinity prediction task.
\item[$\bullet$] \textbf{HGRL-DTA without weighted affinities} (w/o WA) addresses the affinity graph as an unweighted graph, which only considers binary relationships instead of continuous affinities.
\item[$\bullet$] \textbf{HGRL-DTA without message broadcasting} (w/o MB) extracts the global-level representations from the affinity graph, readouts the local-level representations from the molecular graph, and then integrates them using the combination of element-wise addition, element-wise subtraction, and concatenation. Different from HGRL-DTA, this variant learns representations from the affinity graph and the molecular graph separately, without using the global-level affinity information to guide learning local-level molecular properties. Note that this variant keeps the same number of GCN iterations on the molecular graph as HGRL-DTA.
\end{itemize}

\begin{figure}[htbp]
    \centering
    \includegraphics[width=0.7\textwidth]{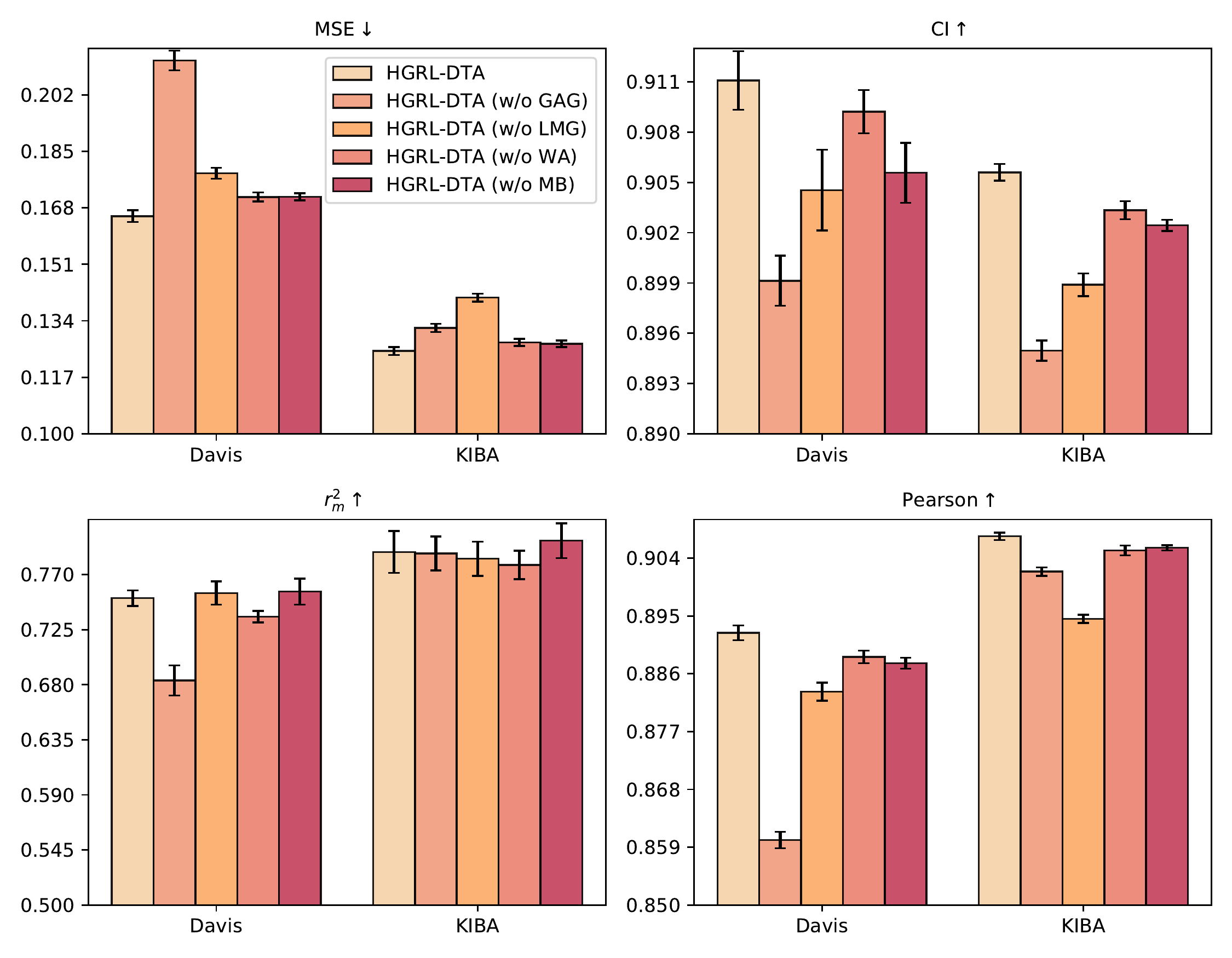}
    \caption{Results of ablation experiments.}
    \label{fig:AblationStudy}
\end{figure}

Figure \ref{fig:AblationStudy} compares HGRL-DTA with its four variants on the two benchmark datasets. Overall, the proposed HGRL-DTA outperforms other variants, which demonstrates the effectiveness of the hierarchical graph learning architecture. In detail, HGRL-DTA (w/o GAG) and HGRL-DTA (w/o LMG) have the most significant performance gaps with HGRL-DTA. These results suggest that global-level and local-level components contribute the most to our model, and removing either component will undermine its predictive performance. Besides, HGRL-DTA (w/o WA) performances worse than HGRL-DTA since it only constructs the affinity graph using binary relationships, which loses more realistic information hidden in continuous affinities. The deprecation of the message broadcasting mechanism in HGRL-DTA (w/o MB) also leads to performance reduction. It indicates that this mechanism is more effective to integrate the global-level affinity information and the local-level molecular properties.

\subsection{Parameter analysis}
To further validate the effectiveness of the similarity-based embedding map for inferring unseen nodes, we analyze the impacts of two major hyper-parameters $simK_d$ and $simK_t$ used in this map.

We conduct the parameter study experiment on the Davis dataset by changing the hyper-parameters $simK_d$ and $simK_t$ from 1 to 8 while keeping other hyper-parameters fixed as default settings. We test $simK_d$ under setting \textbf{S2} where drug is unseen and $simK_t$ under setting \textbf{S3} where target is unknown, respectively. To more directly analyze the effect of inferred representations of nodes absent in the affinity graph, we observe the performance variation of HGRL-DTA w/o LMG, which is a variant of HGRL-DTA only learning global-level representations on the affinity graph for binding affinity predictions.

\begin{figure}[htbp]
    \centering
    \includegraphics[width=1\textwidth]{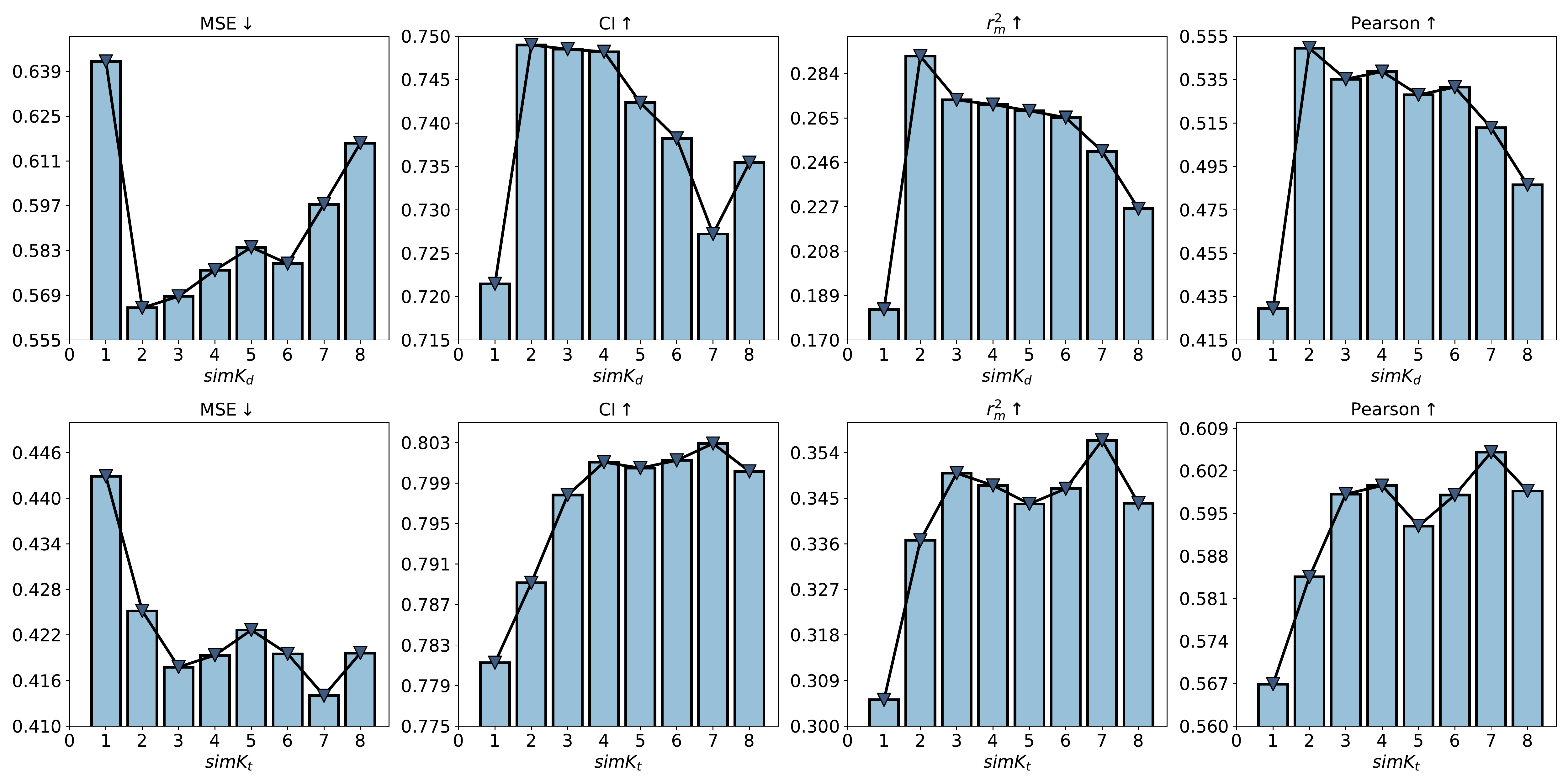}
    \caption{Parameter study of $simK_d$ and $simK_t$ for inferring unseen nodes.}
    \label{fig:ParameterAnalysis}
\end{figure}

As shown in Figure \ref{fig:ParameterAnalysis}, the similarity-based embedding map influences the predictive performance of HGRL-DTA w/o LMG by changing $simK_d$ and $simK_t$. We can see that the model performs best when $simK_d = 2$ and $simK_t = 7$. With the increase of $simK_d$ or $simK_t$, aggregating more known drug/target embeddings to infer unseen drugs/targets can encode more useful information, which leads to dramatic performance improvements. When $simK_d$ or $simK_t$ reaches its optimal value, the performance begins to decline because the aggregation of too many embeddings may introduce redundant and noisy information that can harm the predictive capacity. Furthermore, the non-zero choices of $simK_d$ and $simK_t$ demonstrate the importance of utilizing the similarity-based embedding map to infer unseen nodes in our method.

\subsection{Visualization analysis}
In this subsection, we design an additional experiment to explore the representation power of the proposed model from the view of affinity representations.

To simplify the discussion, we divide affinities into two clusters through predefined thresholds provided in the previous studies \cite{He2017, Tang2014}, where the $pK_d$ value 7 and the KIBA score 12.1 are selected as thresholds for the Davis dataset and the KIBA dataset, respectively. Affinities below the predefined threshold are classified as weak ones and above as strong ones. It should be noted that such division is conducted on the test sets of the two benchmark datasets in setting \textbf{S1}. We preserve the trained HGRL-DTA model in the training phase and then extract the learned representations of testing affinity samples before the final prediction layer.

This experiment analysis is based on an empirical assumption that affinities are expected as close as possible in the same cluster and as far as possible in different clusters in the affinity representation space. In this study, we use Silhouette Coefficient (SC) \cite{Rousseeuw1987}, Calinski-Harabasz Index (CHI) \cite{Calinski1974} and Davies-Bouldin Index (DBI) \cite{Davies1979} to evaluate cluster performance of the affinity representations extracted from various models. Such cluster performance is positively associated with the representation power of the models.

\begin{figure}[htbp]
    \centering
    \includegraphics[width=1\textwidth]{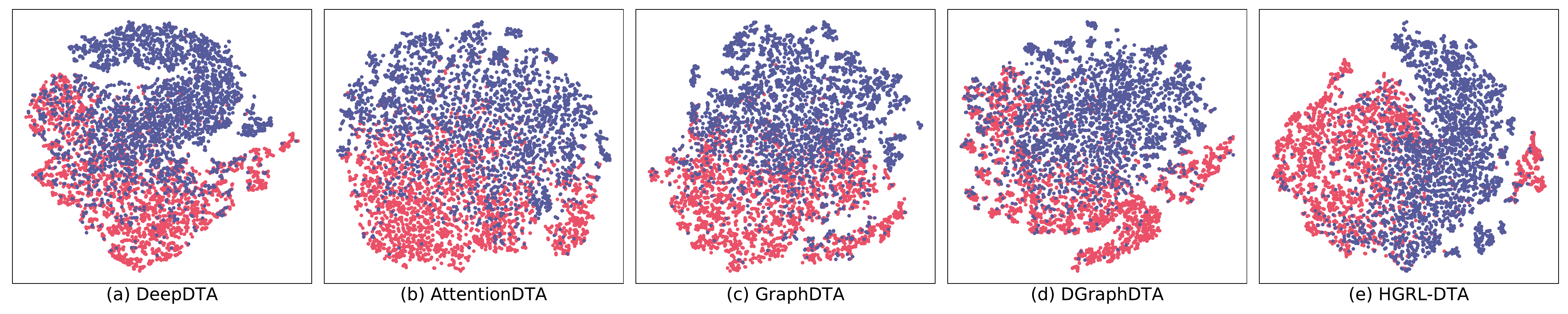}
    \caption{Visualization of affinity representations. Red: weak affinity. Blue: strong affinity.}
    \label{fig:VisualizationAnalysis}
\end{figure}

Table \ref{tab:distances} reports the cluster performance of affinity representations of our model and baselines on the two benchmark datasets. As we can see, our HGRL-DTA model achieves the best cluster performance compared to baseline methods. Moreover, to analyze the affinity representations more intuitively, we sample weak affinities and strong ones with the ratio of 1:1 from the test set of the KIBA dataset and project their representations into 2D space using tSNE \cite{Van2008} for visualization. As illustrated in Figure \ref{fig:VisualizationAnalysis}, HGRL-DTA can well distinguish weak affinities (red) and strong ones (blue); DeepDTA, GraphDTA, and DGraphDTA recognize most of the strong affinities; AttentionDTA differentiates part of affinities. These results indicate that HGRL-DTA allows more delicate affinity representations, which leads to better performance for the binding affinity prediction.

\begin{table}[width=.64\linewidth,pos=htbp]
\caption{Cluster performance of affinity representations.}
\centering
\begin{tabular}{l|lll|lll}
    \toprule
    \multirow{2}{*}{Architecture} & \multicolumn{3}{c|}{Davis} & \multicolumn{3}{c}{KIBA} \\ 
    \cmidrule(lr){2-7}
     & \makecell[c]{SC$\uparrow$} & \makecell[c]{CHI$\uparrow$} & \makecell[c]{DBI$\downarrow$} & \makecell[c]{SC$\uparrow$} & \makecell[c]{CHI$\uparrow$} & \makecell[c]{DBI$\downarrow$} \\ 
    \midrule
    DeepDTA & 0.585 & 3122.789 & 0.730 & 0.305 & 4325.479 & 1.711 \\ 
    AttentionDTA & 0.303 & 592.239 & 1.728 & 0.176 & 2200.669 & 2.470 \\
    GraphDTA & 0.615 & 2751.677 & 0.917 & 0.313 & 5589.479 & 1.593 \\
    DGraphDTA & \bf{0.643} & 2506.929 & 0.906 & 0.353 & 4034.783 & 1.968 \\
	\bf{HGRL-DTA} & 0.639 & \bf{4330.756} & \bf{0.587} & \bf{0.410} & \bf{10385.635} & \bf{1.194} \\
    \bottomrule
\end{tabular}
\label{tab:distances}
\end{table}

\section{Conclusion}
In this paper, we propose a novel hierarchical graph representation learning model to learn the representations of drugs and targets for better drug-target binding affinity prediction. Our model can capture the global-level topological affinities of drug-target pairs and the local-level molecular properties of drugs/targets synergistically, and incorporate such hierarchical graph information using a message broadcasting mechanism. To generalize our model for the cold start situation, we design a similarity-based embedding map to infer the representations of unseen drugs and targets. Extensive experiments under four different scenarios have demonstrated that integrating the topological information of affinity relationships into the representations of drugs and targets can significantly improve the predictive capacity of the models. We also find experimental evidence suggesting that the message broadcasting mechanism is beneficial for the integration of the hierarchical graph information, and the similarity-based embedding map is an effective strategy to infer representations for unseen drugs or targets. In the future, we will extend the proposed method to other biological entity association prediction tasks with hierarchical graph architecture, e.g., drug-drug interaction (DDI) prediction and protein-protein interaction (PPI) prediction.

\section*{Acknowledgements}
This work was supported by the National Natural Science Foundation of China (Grant No.62072206, Grant No.62102158), Huazhong Agricultural University Scientific \& Technological Self-innovation Foundation. The funders have no role in study design, data collection, data analysis, data interpretation, or writing of the manuscript.

\bibliographystyle{model1-num-names}

\bibliography{cas-refs}


\end{document}